\documentclass[letterpaper]{article} % DO NOT CHANGE THIS
\usepackage{aaai24}  % DO NOT CHANGE THIS
\usepackage{times}  % DO NOT CHANGE THIS
\usepackage{helvet}  % DO NOT CHANGE THIS
\usepackage{courier}  % DO NOT CHANGE THIS
\usepackage[hyphens]{url}  % DO NOT CHANGE THIS
\usepackage{graphicx} % DO NOT CHANGE THIS
\usepackage{natbib}  % DO NOT CHANGE THIS AND DO NOT ADD ANY OPTIONS TO IT
\usepackage{caption} % DO NOT CHANGE THIS AND DO NOT ADD ANY OPTIONS TO IT
\usepackage{algorithm}
\usepackage{algorithmic}

\usepackage{newfloat}
\usepackage{listings}
\DeclareCaptionStyle{ruled}{labelfont=normalfont,labelsep=colon,strut=off} % DO NOT CHANGE THIS
\floatstyle{ruled}
\newfloat{listing}{tb}{lst}{}
\floatname{listing}{Listing}
\newcommand{\correspondingauthor}{\thanks{Corresponding Author.}}

\usepackage{multirow}
\usepackage{amsmath}
\usepackage{tcolorbox}
\usepackage{amssymb}
\newcommand{\Xmat}[0]{{{X}}}
\usepackage{booktabs}

\title{PathAsst: A Generative Foundation AI Assistant Towards Artificial General Intelligence of Pathology}
\author{
   Yuxuan Sun\textsuperscript{\rm 1,\rm2,\equalcontrib}, Chenglu Zhu\textsuperscript{\rm 2,\equalcontrib}, Sunyi Zheng\textsuperscript{\rm 2}, Kai Zhang\textsuperscript{\rm 3}, Lin Sun\textsuperscript{\rm 4}, \\Zhongyi Shui\textsuperscript{\rm 1,\rm2}, Yunlong Zhang\textsuperscript{\rm 1,\rm2}, Honglin Li\textsuperscript{\rm 1,\rm2}, Lin Yang\textsuperscript{\rm 2, \correspondingauthor}
}
\affiliations{
    \textsuperscript{\rm 1}College of Computer Science and Technology, Zhejiang University, China\\
    \textsuperscript{\rm 2}Research Center for Industries of the Future and School of Engineering, Westlake University, China\\
    \textsuperscript{\rm 3}Department of Computer Science and Engineering, The Ohio State University, USA\\
    \textsuperscript{\rm 4}School of Computer and Computing Science, Hangzhou City University, China\\
	\{sunyuxuan,yanglin\}@westlake.edu.cn
}

\usepackage{bibentry}
\begin{document}

\maketitle

\begin{abstract}
As advances in large language models (LLMs) and multimodal techniques continue to mature, the development of general-purpose multimodal large language models (MLLMs) has surged, offering significant applications in interpreting natural images. However, the field of pathology has largely remained untapped, particularly in gathering high-quality data and designing comprehensive model frameworks. To bridge the gap in pathology MLLMs, we present PathAsst, a multimodal generative foundation AI assistant to revolutionize diagnostic and predictive analytics in pathology. The development of PathAsst involves three pivotal steps:  data acquisition, CLIP model adaptation, and the training of PathAsst's multimodal generative capabilities. Firstly, we collect over 207K high-quality pathology image-text pairs from authoritative sources. Leveraging the advanced power of ChatGPT, we generate over 180K instruction-following samples. Furthermore, we devise additional instruction-following data specifically tailored for invoking eight pathology-specific sub-models we prepared, allowing the PathAsst to effectively collaborate with these models, enhancing its diagnostic ability. Secondly, by leveraging the collected data, we construct PathCLIP, a pathology-dedicated CLIP, to enhance PathAsst's capabilities in interpreting pathology images. Finally, we integrate PathCLIP with the Vicuna-13b and utilize pathology-specific instruction-tuning data to enhance the multimodal generation capacity of PathAsst and bolster its synergistic interactions with sub-models. The experimental results of PathAsst show the potential of harnessing AI-powered generative foundation model to improve pathology diagnosis and treatment processes. We open-source our dataset, as well as a comprehensive toolkit for extensive pathology data collection and preprocessing at https://github.com/superjamessyx/Generative-Foundation-AI-Assistant-for-Pathology. 
\end{abstract}
\section{Introduction}

In recent years, artificial intelligence has made remarkable strides across various fields~\cite{liu2022investigating,zhuang2021smart}. This is particularly evident in pathology, which has undergone a profound transformation with the introduction of digital pathology and advanced deep learning techniques. The increasing availability of digitized histopathology data, coupled with the exponential growth in the size and complexity of pathology datasets, has necessitated the development of more sophisticated tools to enhance the analytical efficiency of pathologists.

Simultaneously, there has been an upsurge interest in LLMs, with numerous researchers focusing on their development and application. The ultimate goal is to create models with general artificial intelligence capabilities. Among the most prominent examples are OpenAI's ChatGPT and GPT-4. These models have showcased impressive capabilities in human interaction by training through instruction tuning and human feedback, thereby fueling the community's enthusiasm for LLMs.

In the open-source community, LLaMA~\cite{touvron2023llama} has emerged as a compelling model that exhibits performance on par with GPT-3~\cite{brown2020language}, providing promising opportunities for further development. Subsequent models, such as Alpaca~\cite{alpaca} and Vicuna~\cite{vicuna2023}, take advantage of LLaMA and leverage the instruction tuning techniques, enabling them even outperform ChatGPT in certain tasks. Researchers have also explored the realm of multimodal models, creating innovative approaches such as LLaVA~\cite{liu2023visual} and MiniGPT-4~\cite{zhu2023minigpt}. These models demonstrate impressive capabilities in comprehending and interpreting multimodal data, showcasing the advancements in the field.

However, while these advanced MLLMs primarily focus on natural images, the field of pathology faces a notable gap due to the scarcity of high-quality data and limited exploration of model frameworks, which results in a deficiency of pathology-specific MLLMs. In this study, we aim to bridge this gap by exploring both high-quality pathology data collection and the potential application of MLLMs within the pathology domain. We outline our contributions as follows:

\begin{itemize}
	
	\item We gather diverse pathology image-caption pairs from authoritative sources. Through a meticulous process of data cleaning and optimization, we create the PathCap dataset, comprising 207K high-quality samples.
	
	\item We introduce PathCLIP, a pathology-specific CLIP model trained on the PathCap. Compared to prior models, PathCLIP shows superior proficiency in understanding pathology data, achieving state-of-the-art results in pathology image retrieval and zero-shot classification.
	
	\item We integrate PathCLIP and Vicuna-13b to develop PathAsst, a multimodal generative foundational model tailored for pathology. Utilizing the PathCap dataset, we prompt ChatGPT to generate the PathInstruct dataset, which consists of 180K pathology multimodal instruction-following samples. These samples are employed to train PathAsst's generative capabilities. Additionally, we prepare eight pathology-specific sub-models, supplemented with instruction-following data for various scenarios that necessitate sub-model invocation. This equips PathAsst with the ability to discern when to utilize these models for optimal results.
\end{itemize}

\section{Related Work}

\paragraph{Large Language Model (LLM).}
In the early stages, breakthrough models like BERT~\cite{devlin2018bert} and GPT~\cite{radford2018improving}, were introduced, drawing inspiration from the transformer architecture. These models ignited significant interest in the natural language processing (NLP) domain and signaled the beginning of large-scale models in this field. Initially, the full potential of generative models remained largely unexplored.  However, in recent years, as the generative model continue to scale up, more powerful models such as GPT-3~\cite{brown2020language}, T5~\cite{raffel2020exploring}, PaLM~\cite{chowdhery2022palm}, and OPT~\cite{zhang2022opt} are developed. Their emergent abilities~\cite{wei2022emergent} lead these larger models to display markedly superior performance on complex tasks compared to their smaller counterparts.
Furthermore, the introduction of instruction tuning techniques~\cite{ouyang2022training, wang2022benchmarking, wang2022self}, specifically in the realm of LLM, enables the generation of more controllable, practical, and task-specific results. This revolutionary enhancement significantly boosts the zero-shot learning abilities of large models, as exemplified by InstructGPT~\cite{ouyang2022training}, GPT-4~\cite{gpt4}, FLAN-T5~\cite{chung2022scaling}, and FLAN-PaLM~\cite{chung2022scaling}.

\paragraph{Multimodal Large Language Model (MLLM).}
Recent advancements in large-scale multimodal models can be primarily divided into two branches. The first branch is developed based on the LangChain~\cite{Chase_LangChain_2022} approach, where LLM collaborates with various specialized visual models to generate results. Prominent representatives of this branch include Visual ChatGPT~\cite{wu2023visual} and MM-REACT~\cite{yang2023mm}. 
The second branch is implemented by integrating the feature outputs from visual models into the token sequence inputs of the LLM, enabling multimodal generation. This method is represented in models such as BLIP-2~\cite{li2023blip}, 
PaLM-E~\cite{driess2023palm} and Flamingo~\cite{alayrac2022flamingo}. Building upon the instruction-tuning techniques inspired by the LLM community, researchers create multimodal instruction-following datasets to perform MLLM training. This approach promptes the development of models such as LLaVA~\cite{liu2023visual}, MiniGPT-4~\cite{zhu2023minigpt} and LLaMA-Adapter V2~\cite{gao2023llama}. These models demonstrate impressive performance in solving multimodal tasks, as well as advanced multimodal chat capabilities.

\paragraph{Multimodal Model for Pathology.}
While there are numerous applications for multimodal models in natural image analysis, their use in pathological image analysis has been relatively limited to date. The majority of methods employ approaches that combine vision encoder with LSTM~\cite{liu2023describe,zhang2019pathologist,zhang2019text}, yielding fairly satisfactory results. TraP-VQA~\cite{naseem2022vision} is the first attempt to employ vision-language transformer in pathology image processing, which is tested on the PathVQA dataset~\cite{he2020pathvqa} to generate interpretable answers.
More recently, Huang et al.~\cite{huang2023leveraging} compile a large-scale dataset of pathology image-text pairs, sourced from social media platforms such as Twitter. They utilize contrastive vision-language pretraining to establish a foundational model for pathology, demonstrating promising results in pathology zero-shot image-text cross-modal retrieval and zero-shot image classification.

\begin{figure*}[t!]
	\centering  
	\includegraphics[width=0.96\textwidth]{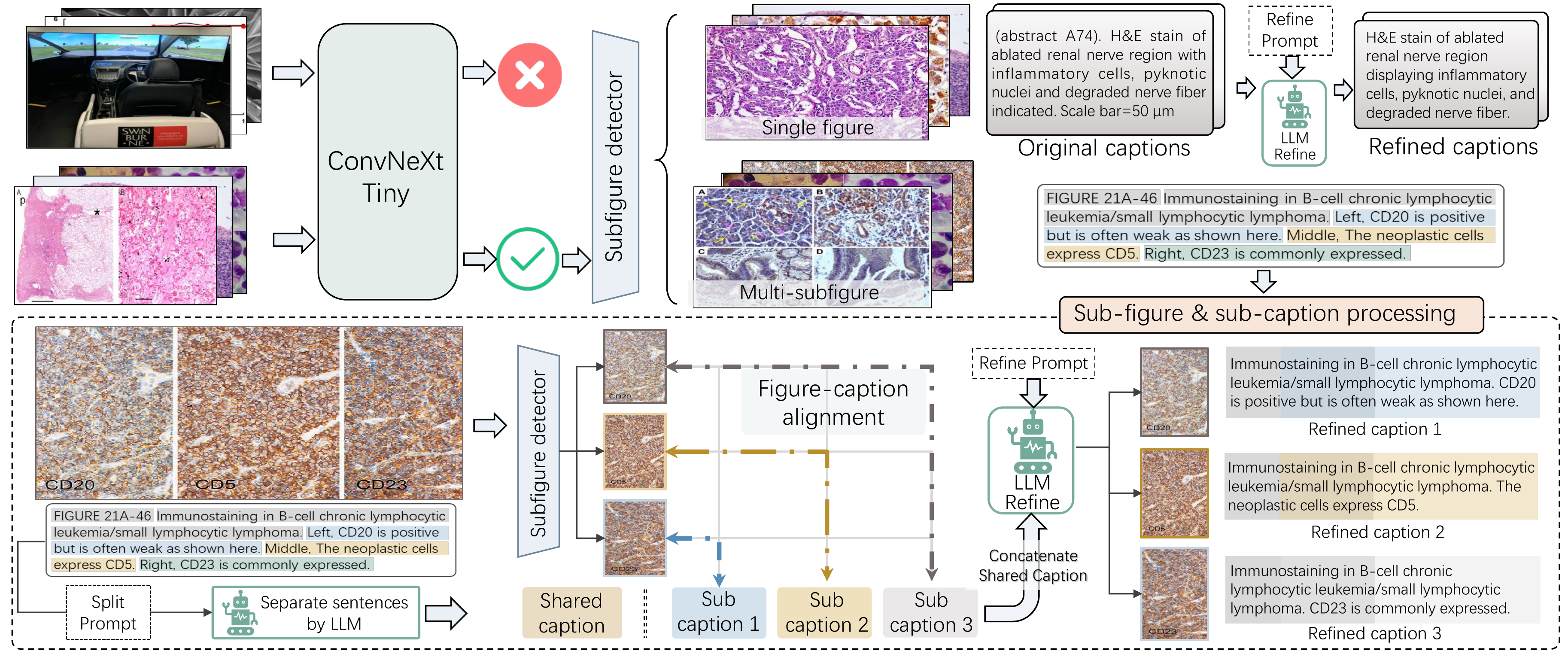}
	\caption{Illustration of data processing: pathology image selection, sub-figure \& caption separation, and refinement.}
	\label{fig:seperate and align}  
\end{figure*}

\begin{figure*}[h!]
	\centering  
	\includegraphics[width=0.95\textwidth]{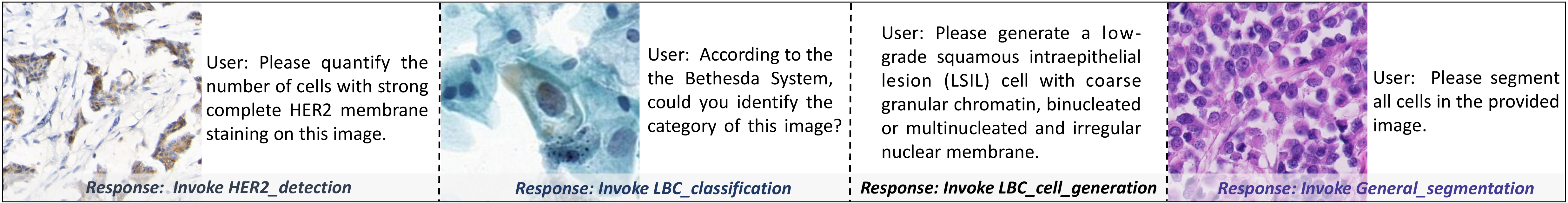}
	\caption{Examples of pathology-specific model-invoking instruction-following samples.}
	\label{fig:pathasst_model_invoking}  
\end{figure*}

\paragraph{Multimodal Datasets.}
Numerous researchers have been dedicating their efforts to contribute valuable datasets that facilitate the advancement of models in the aforementioned domains.  For instance, in the general domain, the community has successfully constructed various datasets, such as CC~\cite{changpinyo2021conceptual} and LAION~\cite{schuhmann2022laion}. In the biomedical field, researchers have released datasets like ROCO~\cite{pelka2018radiology}, MedICAT~\cite{subramanian2020medicat}, and PMC-OA~\cite{lin2023pmc}. In the pathology domain, researchers have recently built the OpenPath~\cite{huang2023leveraging} dataset by crawling Twitter.

Despite significant progress in the field, the domain of MLLM specifically adapted for pathology remains largely untapped. Current models, primarily designed for caption generation, often underperform when compared to specialized professional pathology models. Furthermore, regarding pathology MLLM dataset construction, existing datasets such as ROCO, MedICAT, and PMC-OA are not specifically tailored for this field. The only large-scale dataset, OpenPath, primarily sources its data from Twitter, where the image-text correlation is relatively weak, thus posing challenges for MLLM training. Moreover, the image-text pairs in OpenPath require access to the Twitter API, which carries a significant cost. As a result, there is still a substantial lack of high-quality image-caption datasets in the field of pathology. 
To bridge this gap, we develop two comprehensive pathology multimodal datasets. Building on these datasets, we utilize the power of instruction tuning to significantly improve MLLM's capability in interpreting pathology images.

\begin{figure*}[h]
	\centering  
	\includegraphics[width=\textwidth]{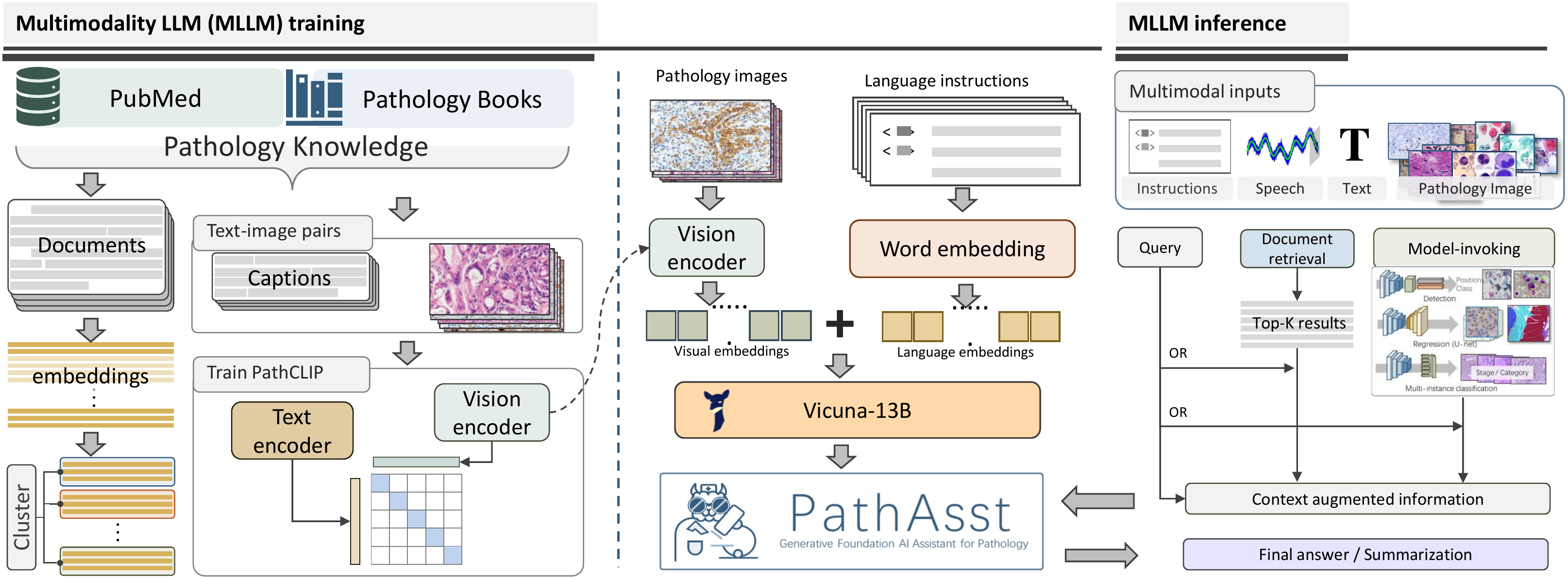}
	\caption{An illustration of the overall framework of PathAsst. The multimodal MLLM training encompasses the training processes of both PathCLIP and PathAsst, as well as the construction of a paper embedding database. The tool-augmented MLLM inference details the process of PathAsst utilizing various tools to enhance the quality of its generated outputs.}
	\label{fig:pathasst_arch}  
\end{figure*}

\section{Pathology Dataset Construction}
\label{Dataset Construction}
In this paper, we propose two datasets tailored for pathology: \textbf{PathCap} and \textbf{PathInstruct}. The PathCap contains 207K high-quality pathology image-caption pairs. Among them, 197K are collected from PubMed and internal pathology guidelines books, while an additional 10K annotations are provided by expert cytologists specializing in liquid-based cytology (LBC). The PathInstruct dataset consists of 180K samples and includes two parts of instruction-following data. The first part is generated by prompting ChatGPT based on curated pathology image-text pairs (refer to step 4 in the subsequent data processing introduction). The second section includes multimodal instruction-following data tailored for model invocation, ensuring the effective use of specialized pathology models based on user intent and image features.

More specifically, data from PubMed are parsed from XML format papers into image-text pairs. For books, we first convert them from PDF to HTML and then parse the content into image-text pairs. Through these efforts, we collect 15M and 2K samples from these respective sources. Although the amount of data available on PubMed is substantial, it should be noted that the proportion of the data related to pathology is limited. Additionally, the clarity of these pathology images is comparatively inferior. Therefore, thorough filtering is required to ensure the quality and relevance of image-text pairs. As shown in Figure \ref{fig:seperate and align}, our data cleansing process is executed methodically, following four carefully designed steps:

\textbf{Step 1: Pathology data selection.} The dataset collected, especially from PubMed, encompasses a wide variety of image sources beyond the scope of pathology. To efficiently select pathology-related data, we manually annotate 20K samples, categorizing them as either pathological or non-pathological. Subsequently, we train a ConvNeXt~\cite{liu2022convnet} model to identify pathological data within the remaining dataset, resulting in a pathology-specific dataset comprising 135K pathology-specific images.

\textbf{Step 2: Sub-figure and sub-caption separation \& alignment.} In many instances, images consist of multiple sub-figures, necessitating precise separation and alignment with their corresponding captions. As depicted in the lower half of Figure \ref{fig:seperate and align}, we address the sub-figure separation by developing a YOLOv7 model~\cite{wang2022yolov7} trained on 2K annotated bounding boxes. Regarding caption separation, conventional rule-based methods often fail to handle the separation of diverse and intricate captions. To overcome this limitation, we leverage the power of ChatGPT to automatically separate approximately 60K captions using carefully crafted prompts. Subsequently, we employ PLIP~\cite{huang2023leveraging} to align sub-image with its corresponding sub-caption by assessing the similarity of visual content and captions. Moreover, we eliminate images with lower resolution, and remove the less relevant image-text pairs, further enhancing the overall quality of the dataset. Ultimately, we acquire 195K high-quality image-text pairs.

\textbf{Step 3: Caption refinement.} As original captions include irrelevant information such as age and disease descriptions, and are not presented in a descriptive style. We design prompts to employ ChatGPT in refining the captions, making them more suitable for training. 

\textbf{Step 4: Instruction-following data generation.} In this step, we select image-text pairs with captions exceeding 12 words. Using these pairs, we produce two types of instruction-following data: detailed description-based and conversation-based. 
The former is created by applying multiple well-designed instructions that inquire about detailed information, 
while the latter involves using ChatGPT to generate conversational Q\&As based on the captions. Additionally, we design special model-invoking instruction-following samples covering a diverse range of scenarios, as depicted in Figure \ref{fig:pathasst_model_invoking}, enabling PathAsst with the capability to appropriately utilize pathology-specific sub-models.

\section{PathAsst Framework Construction}
In this section, we present a comprehensive description of the construction process of PathAsst. This includes the introduction to the design of the model's structure, training methodology, and the tools used for augmented model inference. A general overview can be found in Figure \ref{fig:pathasst_arch}.

\subsection{Model Design and Training}  
PathAsst is designed to integrate the strengths of both the advanced LLM and the CLIP~\cite{radford2021learning} vision encoder to enable enhanced pathological analysis. For the visual component, we employ our custom-trained PathCLIP, complemented by a fully connected (FC) layer. Concerning the LLM component, we utilize Vicuna-13B~\cite{vicuna2023}, a model widely recognized as the closest to ChatGPT in terms of performance. To elaborate, when an input image is provided, it is first encoded into visual tokens via the PathCLIP. Subsequently, the FC  layer maps the image embedding space to the corresponding language embedding space. Finally, both visual and language embeddings are concatenated to the inputs of the  MLLM. In the following, we introduce the detailed training process of PathCLIP and PathAsst.

\paragraph{Training of PathCLIP.}
As one of the core components of PathAsst, the capability of CLIP in interpreting pathological images largely dictates the performance ceiling of PathAsst. Therefore, we develop PathCLIP, a specialized variant of CLIP tailored for pathology. The training process involves fine-tuning a pre-trained OpenAI CLIP base model~\cite{radford2021learning} using our PathCap dataset in a contrastive learning approach, following the training procedure from OpenCLIP repository~\cite{Ilharco_OpenCLIP_2021}. To be specific, for a batch of $N$ image-text pairs, PathCLIP is designed to maximize the cosine similarity between the embeddings of the pathology image and its corresponding text within each batch. Concurrently, it minimizes the cosine similarity amongst the remaining $N^2 - N$ non-pair samples.
This strategy aligns the pathology vision and language space, thereby endowing PathCLIP with a more effective interpretation and analysis of pathology images.

\paragraph{Training of PathAsst.} PathAsst is trained using the PathInstruct dataset through a two-phase training. In the first phase, both the vision encoder and the LLM are frozen, and we only train the FC layer that connects to the vision encoder. This initial phase aims to preliminarily align the vision encoder with the LLM. During this phase, we utilize the detailed description-based part of the PathInstruct. In the second phase, with an aspiration for PathAsst to generate higher-quality and more detailed responses, we extract all the data from books within the PathInstruct dataset, and include samples from PubMed with single images and captions exceeding the length of 50 tokens, resulting in a total training set of 35K samples. Only the PathCLIP is frozen during this phase's training. 

Specifically, we standardize both forms of instruct-following data formats, as shown in Table \ref{train_prompt}. First, we predefine a system message that sets the context for the LLM role. This is followed by a conversation between the user and the assistant, where the user provides instructions, and the assistant responds accordingly based on the instructions. 
To fine-tune our model, we utilize instruction-tuning via next-word prediction. Specifically, the model is trained to optimize the likelihood of generating an accurate response given the input image $\mathcal{I}$ and instruction $\mathbf{X}_{instruct}$. The loss is calculated using the negative log-likelihood of the correct next token in the sequence, with the total loss summed across all time steps, which can be formulated as:
%while maintaining the auto-regressive training objective.
\begin{equation}
	\mathcal{L}(\boldsymbol{\theta}) = -\sum_{t=1}^T \log p\left(x_t \mid \mathcal{I},\mathbf{X}_{instruct}, \mathbf{X}_{\mathrm{a},<t} ; \boldsymbol{\theta}\right),
\end{equation}
Where \(\mathbf{X}_{\mathrm{a},<t}\) refers to the prior tokens in the response sequence, \(\theta\) denotes the trainable parameters of PathAsst. Specifically, during the first phase of training, \(\theta\) corresponds to the parameters of the FC layer. In the subsequent phase, it represents both FC layer and LLM parameters. Meanwhile, \(T\) signifies the length of the ground-truth response, and  $ p\left(x_t \mid \mathcal{I},\mathbf{X}_{instruct}, \mathbf{X}_{\mathrm{a},<t} ; \boldsymbol{\theta}\right) $
represents the probability of generating the \(t\)-th token in the response sequence.

\begin{table}[!t]
	\centering
	\resizebox{0.95\linewidth}{!}{
		\begin{tcolorbox}
			\raggedright
			$\Xmat_{\texttt{system-message}}$  \texttt{<STOP>} $\backslash\texttt{n}$ 
			
			User: $<$image\_token$>$$\backslash \texttt{n}$  \{instruction\} \texttt{<STOP>} $\backslash \texttt{n}$ 
			Assistant: \{response\} \texttt{<STOP>} $\backslash \texttt{n}$  \\
			
			User: \{instruction\} \texttt{<STOP>}$\backslash\texttt{n}$\\
		 	Assistant:  \{response\} \texttt{<STOP>} $\backslash \texttt{n}$  ......
	\end{tcolorbox}}
	
	\caption{Illustration of instruction-following data format, where \{instruction\} represents the user query, \{response\} denotes the corresponding answer.  The $\Xmat_{\texttt{system-message}}$ is set as: A dialogue between a professional pathology assistant and a human. The assistant provide informative, helpful, and detailed answers. The $<$STOP$>$ is represented by \#\#\#, while $<$image\_token$>$ stands for the tokens corresponding to the image tokens. During the model training, only \{response\} is considered when calculating loss.}
	\label{train_prompt}
\end{table}

\subsection{Tool Augmented MLLM Inference}  
To augment PathAsst's capabilities and offer more precise responses, we prepare two types of tools that PathAsst can employ during its inference phase. One leverages pathology-specific computer vision (CV) sub-models, while the other focuses on paper retrieval. These tools not only enrich the context for PathAsst but also enable tasks beyond text generation, such as image generation and segmentation.

\paragraph{Pathology-specific CV Model Zoo.} We integrate eight specialized pathological models into PathAsst for seamless invocation: \textbf{(1)} LBC (liquid-based cytology) classification model: This model is based on ConvNeXt-Tiny~\cite{liu2022convnet}, specifically designed for liquid-based cervical cytology image classification. Through the analysis of abnormal cell morphologies within the image, it effectively classifies the image into one of the six categories as defined by The Bethesda System (TBS). \textbf{(2)} LBC detection model: We utilize YOLOv7~\cite{wang2022yolov7} as the backbone for developing our detection model, which is employed to identify abnormal cells within image patches. This model is specifically designed to detect the five classes of non-normal cells as defined in TBS. \textbf{(3)} Hematological cell detection model: This model, developed based on YOLOv7, specializes in blood cell classification, which is crucial for diagnosing various hematological conditions. \textbf{(4)} LBC cell generation model: This model is developed based on Stable Diffusion~\cite{rombach2022high}, which is capable of generating specific cells based on user input, such as `generate an image of a cell with nuclei enlarged 2-2.5 times'. \textbf{(5)} HER2 detection model, \textbf{(6)} PD-L1 detection model and \textbf{(7)} Ki67 detection model are developed using DPA-P2PNet~\cite{shui2023deformable} for immunohistochemical cell detection and classification. \textbf{(8)} General segmentation model: 
Benefiting from the outstanding general segmentation quality of the Segment Anything Model~\cite{kirillov2023segment}, we directly employ it as our pathology image segmentation model. 

Once PathAsst invokes a particular specialized model, it processes both the user's query and the output of the invoked model to formulate a conclusive response, resulting in a more precise and effective interaction with the user.

\begin{figure*}[h!]
	\centering  
	\includegraphics[width=\linewidth]{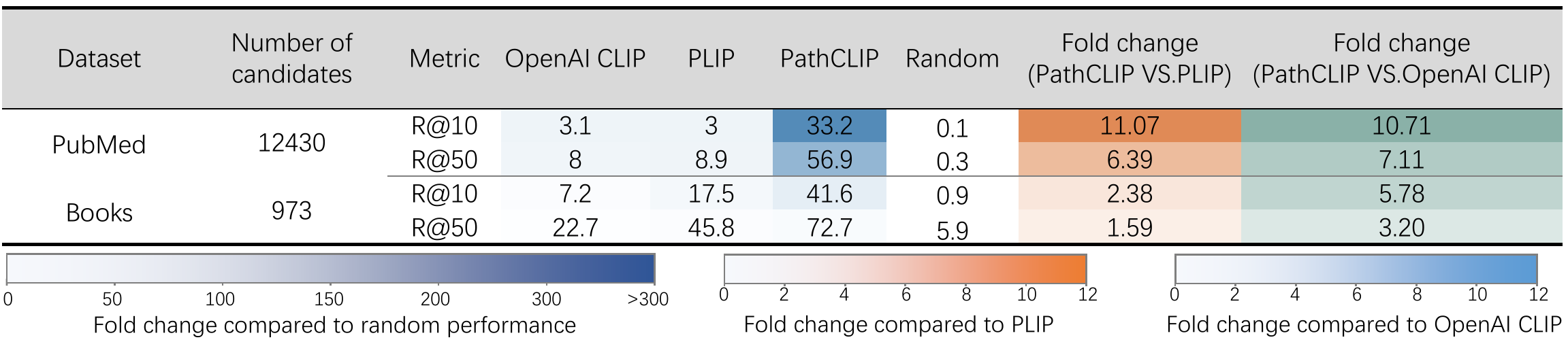}
	\caption{Comparative assessment of image retrieval performance between CLIP models across collected datasets.}
	\label{fig:clip_retrieval}  
\end{figure*}

\paragraph{Enhancing Responses through Paper Retrieval.} In the realm of pathology, even the highly recognized GPT-4 struggles with specific queries that necessitate deep domain knowledge, especially apparent when addressing questions involving the most recent research. Taking inspiration from Langchain's approach~\cite{Chase_LangChain_2022} for building local knowledge databases, we gather 5.3M article abstracts from PubMed. We utilize PubMedBERT \cite{gu2021domain} for abstract embedding extraction and Faiss \cite{johnson2019billion} for the efficient storage of these embeddings. To expedite inference efficiency, a preliminary abstract clustering is conducted. Upon user query, our system allows the extraction of relevant information from this paper database, serving as context information to amplify the precision of LLM's responses.

\begin{figure}[h!]
	\centering
	\includegraphics[width=\linewidth]{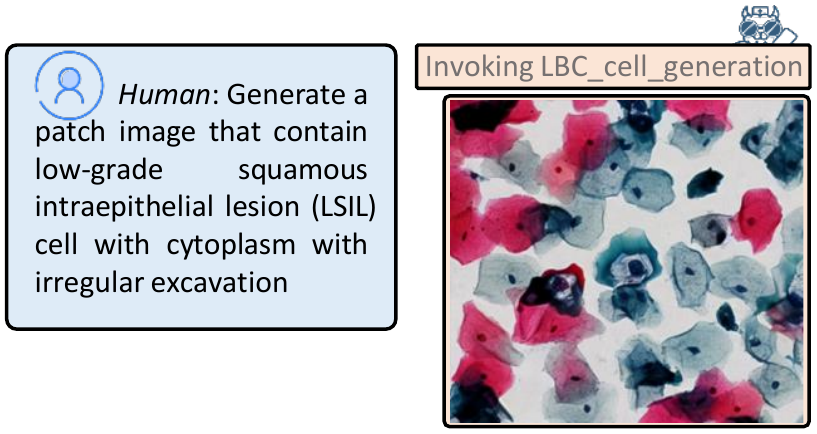}
	\caption{Example of PathAsst calls generation model.}
	\label{fig:pathasst_generation}  
\end{figure}

\begin{figure*}[h!]
	\centering  
	\includegraphics[width=\linewidth]{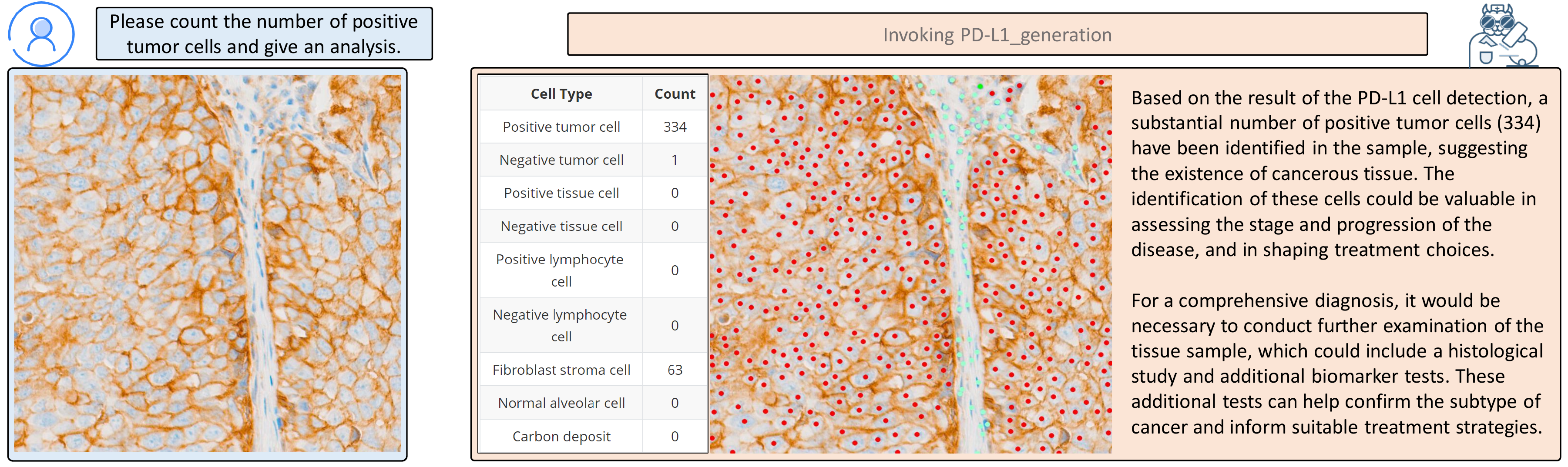}
	\caption{An example of PathAsst invokes the PD-L1 detection model for assistance.}
	\label{fig:pathasst_detection}  
\end{figure*}

\begin{table}[t!]
	\centering
	\resizebox{\linewidth}{!}{
		\begin{tabular}{@{}ccccc@{}}
			\toprule
			Model       & CRC & WSSS4LUAD & LC-lung & LC-colon \\ \midrule
			OpenAI CLIP & 22.2              & 61.6   & 31.5              & 75.7    \\
			PLIP        & 53.1              & 69.5  & 86.0              & 87.0    \\
			PathCLIP    & \textbf{54.2}              & \textbf{81.1}   & \textbf{88.7}             & \textbf{94.3}   \\ \bottomrule
	\end{tabular}}
	%	\vspace{-1mm}
	\caption{Comparative evaluation of zero-shot image classification performance across different CLIP models.}
	\label{tab:clip_zeroshot classfication}  
\end{table}

\begin{figure*}[h!]
	\centering  
	\includegraphics[width=\linewidth]{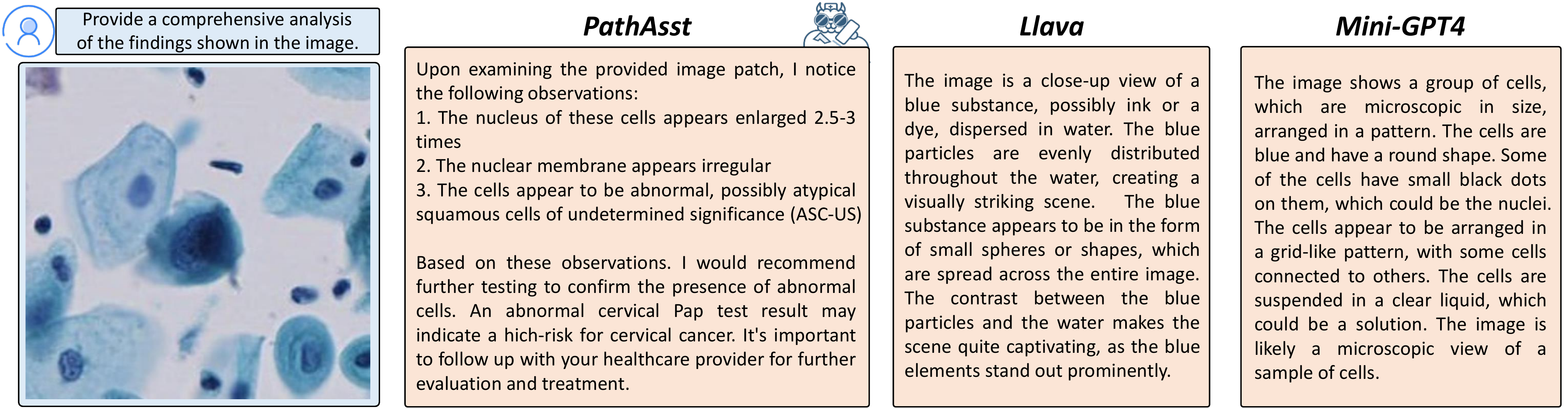}
	\caption{An example of PathAsst, LLaVA, and MiniGPT-4's capability in  interpreting pathology images.}
	\label{fig:pathasst_interpret}  
	\vspace{-2mm}
\end{figure*}

\section{Experiments}
\paragraph{Evaluation Datasets Construction.}
We construct and gather a series of test datasets to evaluate the performance of the proposed PathCLIP and PathAsst.

For the evaluation of zero-shot classification of PathCLIP, we collect: (1) CRC100K dataset~\cite{kather_jakob_nikolas_2018_1214456}: This is a collection of 100K image patches derived from H\&E stained histological images of both colorectal cancer and normal tissue, categorized into nine tissue classes, including Adipose, Background, Debris, Lymphocytes, Mucus, Smooth Muscle, Normal Colon Mucosa, Cancer-Associated Stroma, and Colorectal Adenocarcinoma Epithelium. (2) WSSS4LUAD~\cite{han2022wsss4luad}: This dataset comprises patch-level annotations from 87 whole slide images. In this case, we focus on the tumor and normal classes, which yields a total of 6,579 tumor and 1,832 normal images. In order to assess the model on these datasets, labels are transformed into complete sentences. For instance, the label `tumor' is rephrased as `A H\&E image of a tumor.' (3) LC25000~\cite{borkowski2019lung}. This dataset comprises tissue samples from lung and colon adenocarcinomas, divided into two distinct subsets: the LC-lung and the LC-colon. The LC-lung encompasses 15,000 images and includes classifications of lung adenocarcinomas, lung squamous cell carcinomas, and benign lung tissues. On the other hand, the LC-colon subset, containing 10,000 images, is categorized into colon adenocarcinomas and benign colonic tissues. The F1 score is used as the metric for evaluation.

For the cross-modal retrieval validation of PathCLIP, we employ the test set from the PubMed section of our collected data, along with data from books. Note that the data from books are not included during the training phase of the PathCLIP, hence providing an evaluation in the context of unseen domains. Given that the length of some captions in these datasets is relatively long and may exceed the token length limitation of CLIP, we opt for samples with captions that are fewer than 77 tokens.
The $R@k$ metric is used to assess the performance of image retrieval, which measures whether the correct image is presented among the $Topk$ retrieved images.  

For the validation of PathAsst, we employ the PathVQA dataset~\cite{he2020pathvqa}, which comprises 32,799 questions derived from 4,998 pathology images. The type of questions includes open-ended questions typically beginning with what, where, and when, as well as close-ended questions requiring yes/no responses. We measure model performance of close-ended questions using accuracy, and evaluate open-ended questions with F1-score. 

\begin{table}[t!]
	\centering
	\resizebox{\linewidth}{!}{
		\begin{tabular}{@{}ccc@{}}
			\toprule
			& \multicolumn{2}{c}{\textbf{PathVQA}} \\ \cmidrule(l){2-3} 
			\multirow{-2}{*}{\textbf{Method}} & Closed & Open \\ 
			\midrule
			M2I2~\cite{li2022self} & 88.0 & 36.3 \\
			CLIP-ViT w/ GPT2~\cite{van2023open} & 87.0 & \textbf{40.0} \\
			MMQ~\cite{do2021multiple} & 84.0 & 13.4 \\ 
			\midrule
			LLaVA~\cite{liu2023visual} & 81.0 & 19.2 \\
			BLIP-2 Flan-T5 XXL~\cite{li2023blip} & 80.1 & 34.1 \\
			PathAsst (w/ CLIP) & 89.7 & 37.6 \\
			PathAsst (w/ PathCLIP) & \textbf{90.9} & 38.4 \\ 
			\bottomrule
	\end{tabular}}
	\caption{Comparison of various methods on PathVQA.}
\end{table}

\paragraph{Statistical Results.} As shown in Table~\ref{tab:clip_zeroshot classfication} and Figure \ref{fig:clip_retrieval}. Our analysis demonstrates that our PathCLIP significantly surpasses the baseline OpenAI CLIP model, consistently outperforming the state-of-the-art (SOTA) pathology model, PLIP, in tasks such as cross-modal image retrieval and zero-shot image classification. To be specific, PathCLIP achieves a remarkable improvement in the $R@10$ retrieval on the PubMed dataset, with a 10.71-fold and 11.07-fold increase compared to the OpenAI CLIP and PLIP models, respectively. In the context of unseen domain data, the retrieval $R@10$ on the books is 5.78 times and 2.38 times that of OpenAI CLIP and PLIP, respectively. Considering the zero-shot classification tasks, PathCLIP achieves a substantial improvement in F1-score compared to CLIP, with notable gains of 32\%, 19.5\%, 57.2\%, and 18.6\% on the CRC100K, WSSS4LUAD, LC-lung, and LC-colon datasets, respectively. Furthermore, even when compared to the previous SOTA PLIP model, PathCLIP shows an increase of 1.1\%, 11.6\%, 2.7\%, and 7.3\% on these datasets, respectively. For the evaluation on PathVQA, PathAsst significantly outperforms the prior MLLM model in both closed-form and open-ended question types. Specifically, it surpasses LLaVA by 8.7\% and 18.4\% in these two question types, respectively. This underscores the importance of training with PathInstruct data. Further enhancements of 1.2\% and 0.8\% are noted after substituting CLIP with PathCLIP, indicating that the incorporation of PathCLIP enhances PathAsst's understanding of pathology images. Compared with the performance of the previous SOTA model, which directly extracts the statistical number from their reports, PathAsst achieves considerable improvements in closed-ended questions, although it slightly underperforms the SOTA model in open-ended questions.

\paragraph{Demonstration Showcase of PathAsst.} Here, we showcase several examples of PathAsst's robust capabilities in handling complex pathology tasks. As shown in Figure \ref{fig:pathasst_generation}, PathAsst is capable of recognizing the user's need to generate an LBC cell that belongs to the LSIL category with irregular excavated cytoplasm. It accomplishes this by invoking the LBC cell generation model. This advanced functionality empowers users to create a diverse range of LBC cells that are precisely tailored to their specific needs. 

Figure \ref{fig:pathasst_detection} illustrates another example of PathAsst employing a model invocation, where the user requires to count the positive cells in the image, which can be challenging through direct multimodal generation. Therefore, PathAsst chooses to invoke the PD-L1 cell detection model. It automatically marks the predicted points on the cells in the image and provides the statistical results for further analysis with LLM. In this case, LLM generates a markdown-formatted table to display the results along with the corresponding analysis. 

Furthermore, Figure~\ref{fig:pathasst_interpret} demonstrates PathAsst's ability to interpret pathology images independently. In comparison to LLaVA and MiniGPT-4, PathAsst places greater emphasis on cell morphology and features, such as enlarged nucleus and irregular nuclear membrane. In contrast, LLaVA fails to recognize the image as pathological, while  MiniGPT-4 generates simplistic descriptions such as `cells are blue and have a round shape' and `cells are suspended in a clear liquid.'

\section{Conclusion}
In this study, we construct PathCap and PathInstruct datasets, comprising 207K pathology image-text pairs and 180K instruction-following samples, by systematically collecting and processing pathology data from various sources. Leveraging these high-quality datasets, we propose PathCLIP and PathAsst. PathCLIP exhibits powerful capabilities in pathology cross-modal retrieval and zero-shot classification. PathAsst, an instruction-tuned foundation model, is a synergy of the powerful vision encoder PathCLIP and the Vicuna-13b LLM, equipped with an established toolkit that includes eight pathology-specific models and a 5.3 million-sized paper retrieval system. PathAsst not only showcases impressive capabilities in pathology multimodal dialogue and interpreting pathology images, but also the ability to handle more complex pathology tasks by the invocation of these established pathology tools. We hope that the construction of model frameworks and datasets can offer insights and aid in the advancement of pathology foundational models.

\section{Acknowledgements}
This study was partially supported by the National Natural Science Foundation of China (Grant No.92270108), Zhejiang Provincial Natural Science Foundation of China (Grant No.XHD23F0201), and the Research Center for Industries of the Future (RCIF) at Westlake University.

\bibliography{aaai24}

\end{document}